\documentclass{article}

\PassOptionsToPackage{numbers, compress}{natbib}

\usepackage[final]{neurips_2020}

\usepackage[utf8]{inputenc} 
\usepackage[T1]{fontenc}    
\usepackage{hyperref}       
\usepackage{url}            
\usepackage{booktabs}       
\usepackage{amsfonts}       
\usepackage{nicefrac}       
\usepackage{microtype}      
\usepackage{amsmath}
\usepackage{graphicx}
\usepackage{algorithm2e}
\usepackage{multirow}
\usepackage{caption}
\usepackage{subcaption}
\usepackage{cancel}
\usepackage{changes}
\usepackage{booktabs}

\setdeletedmarkup{\cancel{#1}}

\hypersetup{
	colorlinks=true,
	linkcolor=red,
	filecolor=blue,
	urlcolor=blue,
}

\title{Off-Policy Self-Critical Training for Transformer in Visual Paragraph Generation}

\author{Shiyang Yan, Yang Hua, Neil M. Robertson \\
elyotyan@gmail.com \\
}

\begin{document}
	
	\maketitle
	
	\begin{abstract}
		Recently, several approaches have been proposed to solve language generation problems. Transformer is currently state-of-the-art seq-to-seq model in language generation. Reinforcement Learning (RL) is useful in solving exposure bias and the optimisation on non-differentiable metrics in seq-to-seq language learning. However, Transformer is hard to combine with RL as the costly computing resource is required for sampling. We tackle this problem by proposing an off-policy RL learning algorithm where a behaviour policy represented by GRUs performs the sampling. We reduce the high variance of importance sampling (IS) by applying the truncated relative importance sampling (TRIS) technique and Kullback-Leibler (KL)-control concept. TRIS is a simple yet effective technique, and there is a theoretical proof that KL-control helps to reduce the variance of IS. We formulate this off-policy RL based on self-critical sequence training. Specifically, we use a Transformer-based captioning model as the target policy and use an image-guided language auto-encoder as the behaviour policy to explore the environment. The proposed algorithm achieves state-of-the-art performance on the visual paragraph generation and improved results on image captioning.
	\end{abstract}
	
	\section{Introduction}
	Transformer (self-attention) is a kind of seq-to-seq models, which shows breakthrough successes in natural language processing (NLP), such as machine translation and image captioning~\cite{vaswani2017attention}~\cite{cornia2019m}. Seq-to-seq models are usually trained using either Maximum Likelihood Estimation (MLE) or Reinforcement Learning (RL)~\cite{yu2017seqgan}~\cite{rennie2017self}. Especially, RL for seq-to-seq models can tackle two problems in language generation: (1). The exposure bias, referring to the train-test discrepancy in seq-to-seq models. The training uses the ground-truths while the testing generates a new token based on the previously generated ones. (2). The gradient estimation towards optimisation for non-differentiable evaluation metrics such as BLEU or CIDEr~\cite{rennie2017self}. Indeed, RL has brought significant performance gain in image captioning and language generation~\cite{rennie2017self}~\cite{yu2017seqgan}. However, there is less literature on Transformer performing RL~\cite{parisotto2019stabilizing}. On-policy RL is known to be sample inefficient, and this is especially serious for Transformer in visual paragraph generation where the generated paragraph usually contains about 200 words or more~\cite{krause2017hierarchical}. Expensive computing resource is required for the gradient graph of the decoder, which is established in each time step for on-policy training, making the training even in-feasible.
	
	Off-policy RL, on the contrary, is to use another independent behaviour policy to explore the environment and transfer the experience to the target policy. Off-policy is sample efficient~\cite{gu2016q} and also can largely reduce the computing resources required. In RL, the concept of off-policy is usually rooted in value-based RL~\cite{mnih2013playing}~\cite{munos2016safe}. However, the RL in NLP is usually policy-based RL learning methods, e.g., REINFORCE-like~\cite{williams1992simple} algorithms~\cite{yu2017seqgan}~\cite{rennie2017self} and actor-critic~\cite{bahdanau2016actor} as the action space (vocabulary) is large. The value-based RL is not advantageous in dealing with large action space~\cite{keneshloo2019deep}.

	Also, off-policy RL is sometimes inaccurate as there exists a discrepancy between the target and the behaviour policy. A true off-policy where the target and behaviour policy is non-correlated is extremely hard~\cite{fujimoto2018off}. The well-known off-policy RL learning algorithms such as DQN~\cite{mnih2013playing} and DDPG~\cite{lillicrap2015continuous}, are only capable of learning with data correlated to their current policy~\cite{fujimoto2018off}. A common way of approximation in off-policy is using Importance Sampling (IS) estimators~\cite{farajtabar2018more}~\cite{jiang2015doubly}, which tries to correct the mismatch in the distributions under the behaviour policy and target policy. IS, however, has high variance when the two policy distributions are very different. The ratio of the two probabilities sampled becomes either small or large (sometimes infinite), which leads to huge variance. This phenomenon is noticeable when the episode of RL is long, like when dealing with long sentence generation in visual paragraph generation.

	Hence, we propose an off-policy self-critical sequence training based on its on-policy version~\cite{rennie2017self}, a REINFORCE-like Policy Gradient algorithm and apply it for visual paragraph generation. We employ the smooth version of IS, i.e. truncated relative importance sampling (TRIS)~\cite{humayoo2018relative} to reduce the variance of the conventional IS. TRIS is proved to be effective in reducing the variance of IS as it introduces a relative distribution ratio, which is bounded. Also, there is evidence that the Kullback-Leibler (KL) divergence between the target and behaviour policy influence the variance of IS~\cite{wexler2012importance}. KL-control studies an RL problem in which the agent tries to maximise the task-related reward while minimising deviation from a prior policy (behaviour policy). Consequently, when training the target policy with off-policy RL, we penalise its divergence from the behaviour policy with KL-control~\cite{rawlik2013stochastic}. We add a term of the KL divergence between the target policy and the behaviour policy in the value function of our RL and incorporate it into the self-critical sequence training.

	To be specific, we train Meshed Transformer~\cite{cornia2019m} optimised under the proposed off-policy self-critical sequence training for visual paragraph generation. We design a GRU-based image-guided language auto-encoder as the behaviour policy and treat our Transformer as the target policy. The target policy will learn self-critical rewards while minimising the divergence from the behaviour policy, reducing the variance of the TRIS. To summarise, our contributions are threefold: (1) We propose a novel off-policy self-critical sequence training framework, making the RL learning feasible for Transformer. (2) We reduce the variance of the IS ratio, which is in off-policy RL approximation, by applying TRIS and the concept and techniques of KL-control. (3) We achieve state-of-the-art results on visual paragraph generation and improved results on image captioning. Empirical evidence also shows that the IS variance can be significantly reduced.
	
	\section{Related Works}
	\subsection{Off-Policy RL Learning}
	RL with replay buffer~\cite{lin1992self} can be considered as a standard tool for off-policy learning~\cite{mnih2015human}. In these schemes, the behaviour policy in the replay buffer is somehow related to the target policy~\cite{mnih2013playing}, which is not a `true' off-policy RL learning~\cite{fujimoto2018off}. For example, Isele et al.~\cite{isele2018selective} see that the performance of an agent is most reliable when the distribution of data in the replay buffer matched the test distribution.
	
	Many approaches~\cite{dudik2011doubly}~\cite{farajtabar2018more} use IS to re-weight the probability distribution when the target policy is different from behaviour policy. However, IS is with high variance, preventing the model from achieving stable performance. Hanna et al.~\cite{hanna2018importance} use function approximation to estimate the behaviour policy to reduce the variance. Liu et al.~\cite{liu2018breaking} models the stationary state visiting distribution of the behaviour policy for infinite-horizon off-policy RL tasks. Humayoo et al.~\cite{humayoo2018relative} apply a simple technique, TRIS, to solve this problem.
	
	KL-control is a branch of stochastic optimal control, where the KL divergence from other distributions is applied in regularisation~\cite{abdolmaleki2018maximum}~\cite{jaques2019way}.  An example in the on-policy policy gradient is Trust Region Policy Optimisation (TRPO)~\cite{schulman2015trust}, where a KL penalty term is incorporated in the value function of the Policy Gradient algorithm. KL-control has also been used to improve transfer learning between MLE training on data and training with RL~\cite{jaques2017sequence}.
	

	\subsection{Image Paragraph Generation}
	
	Regions-Hierarchical~\cite{krause2017hierarchical} introduces the first large-scale paragraph captioning dataset, which utilises the images from Visual Genome dataset and adds new annotations. The dataset shows more pronouns, verbs and more diversities than the single sentence captioning dataset, which is more challenging.
	
	Approaches~\cite{krause2017hierarchical}~\cite{liang2017recurrent}~\cite{chatterjee2018diverse} propose different types of hierarchical model structures to generate the visual paragraphs, with an effective coupling mechanism between sentences within one paragraph. These hierarchical architectures model each sentence and couple the sentences into one paragraph, often with more superior performance than the flat models~\cite{krause2017hierarchical}. Advanced methods like VAE~\cite{chatterjee2018diverse}, GAN~\cite{liang2017recurrent} are applied to boost the performance further. However, we see less literature on Transformer-like models under RL for visual paragraph generation as the sampling in on-line RL is computing-expensive.
	

	\section{Methods}
	
	In this section, we first formulate the RL setting of Transformer in visual paragraph generation, then introduce our off-policy self-critical framework for optimisation.
	\subsection{Formulation of Visual Paragraph Generation in On-Policy Self-Critical Training}
	We consider the visual paragraph generation process as a finite Markov Decision Process (MDP). Transformer can be viewed as an agent, which interacts with the environment (words and image features). In the MDP $\{S, A, P, R, \gamma\}$, $S=\{s_0, ..., s_T\}$ is the state space, $A = \{a_0, ..., a_T\}$ is an action space. $P(s_{t+1}|s_t, a_t)$ is the state transition probability, $R(s_t, a_t)$ is the reward function and $\gamma \in (0,1] $ is the discount factor. The agent selects an action, from a conditional probability distribution, which is called the policy $\pi_{\theta}(a|s)$, parametrised by $\theta$. In visual paragraph generation, the state space composed of image features ($I_F$) and actions generated so far, described as $s_t = \{I_F, a_0, a_1, a_2, ..., a_{t-1}\}$. Value functions are the expectation of  accumulative discounted future reward, measuring how good each state is. There are two kinds of value functions: the state value function $V^{\pi}(s_t)$ and the state-action value function $Q^{\pi}(s_t, a_t)$, which are defined as follows:
	\begin{equation}
	\begin{split}
	& V^{\pi}(s_t) = \mathbb{E}_{a_t, s_{t+1}, ... \sim \pi} \Big [\sum_{l=0}^{T} \gamma^{l} r_{t+l} | S= s_{t} \Big]  \\
	& Q^{\pi}(s_t, a_t) = \mathbb{E}_{s_{t+1}, a_{t+1}, ... \sim \pi} \Big [\sum_{l=0}^{T} \gamma^{l} r_{t+l} | S_t= s_{t}, A_t = a_{t} \Big].    \\
	\end{split}
	\end{equation}
	
	The agent tries to maximise the accumulative reward and update the parameters, the loss function is expressed as follows:
	\begin{equation}
	L(\theta) = V^{\pi}(s_0) =\mathbb{E}_{\pi} \Big [\sum_{l=1}^{T} \gamma^{t-1} r_{t} \Big].
	\end{equation}
	For Policy Gradient methods~\cite{sutton2000policy}, which are widely applied in sequence generation problems, the optimisation can be formulated as:
	\begin{equation}
	\nabla_{\theta}L(\theta) = \mathbb{E}_{\pi} \Big [ Q^{\pi}(s_t, a_t) \nabla_{\theta} log\pi_{\theta}(a_t|s_t) \Big].
	\end{equation}
	The Policy Gradient is unbiased, but with high variance. A common way to address this issue is using an arbitrary baseline $b(s_t)$, which is described as follows:
	\begin{equation}
	\nabla_{\theta}L(\theta) = \mathbb{E}_{\pi} \Big [ (Q^{\pi}(s_t, a_t) - b(s_t)) \nabla_{\theta} log\pi_{\theta}(a_t|s_t) \Big].
	\end{equation}
	The baseline is an arbitrary function, which should be independent from the action $a_t$. The Q function appears in the above equations in self-critical sequence learning is set as the expectation of the accumulated rewards. As there is no intermediate reward in language generation task, the self-critical training uses a single sample from Monte Carlo sampling to approximate the Q function, which, in reality, is the CIDEr score of the sampled sentence $\text{CIDEr}^s$. The self-critical uses the baseline CIDEr score $\text{CIDEr}\textsuperscript{$\wedge$}$ from greedy sampling to reduce the variance in Policy Gradient,
	\begin{equation}
	\nabla_{\theta}L(\theta) = \mathbb{E}_{\pi} \Big [ (\text{CIDEr}^s - \text{CIDEr}\textsuperscript{$\wedge$}) \nabla_{\theta} log\pi_{\theta}(a_t|s_t) \Big].
	\end{equation}

	\subsection{The Proposed Off-Policy Self-Critical Training}
	One reason for the instability of off-policy learning is the discrepancy between distributions of the target and behaviour policies as we wish to gather data from the distributions of target policy but sample data from the distribution of the behaviour policy.

	\paragraph{Importance Sampling (IS).}
	IS~\cite{huber1981wiley}~\cite{precup2000eligibility} is a classical approach in handling the discrepancy between the target and behaviour policies. If the behaviour policy is $
	\pi^b$, if $\tau=\{a_1, ..., a_{t}, ...,  a_{T}\}$, then IS in off-policy self-critical learning can be written as:
	\begin{equation}
	\begin{split}
	&\nabla_{\theta}L(\theta) 
	= \mathbb{E}_{\pi^b} \Big [(\text{CIDEr}^s - \text{CIDEr}\textsuperscript{$\wedge$}) \frac{\pi(\tau)}{\pi^b(\tau)}
	\nabla_{\theta} log\pi_{\theta}(a_t|s_t)   \Big] \\
	& = \mathbb{E}_{\pi^b} \Big [ (\text{CIDEr}^s - \text{CIDEr}\textsuperscript{$\wedge$}) (\prod_{t=0}^T{\mu_t}) \nabla_{\theta} log\pi_{\theta}(a_t|s_t)  \Big], \\
	\end{split}
	\end{equation}
	where $\mu_t = \frac{\pi_\theta(a_t|s_t)}{\pi^b(a_t|s_t)}$ is the importance ratio. The IS is with high variance, especially when the discrepancy between the distributions of the target policy and the behaviour policy is large, as the ratio of the probability becomes unstable.
	
	\paragraph{Truncated Relative Importance Sampling (TRIS).}
	Relative Importance Sampling (RIS)~\cite{sugiyama2015introduction}~\cite{yamada2013relative}~\cite{humayoo2018relative} can be applied to smooth the IS so as to reduce the variance, which is described as follows:
	\begin{equation}
	\begin{split}
	& \mu_t^r = \frac{ \pi_\theta(a_t|s_t)}{\lambda { \pi_\theta(a_t|s_t)} + (1 - \lambda){ \pi^b(a_t|s_t)}}  \\
	&  \nabla_{\theta}L(\theta) = \mathbb{E}_{\pi^b} \Big [ (\text{CIDEr}^s - \text{CIDEr}\textsuperscript{$\wedge$}) (\prod_{t=0}^T{\mu_t^r}) \nabla_{\theta} log\pi_{\theta}(a_t|s_t)  \Big],\\
	\end{split}
	\end{equation}
	where the RIS is bounded as it is no greater than $\frac{1}{\lambda}$, as proved in~\cite{humayoo2018relative}. Accordingly, RIS has bounded variance and is also with low bias~\cite{sugiyama2015introduction}. The probability ratio $\prod_{t=0}^T{\mu_t^r}$ does not involve a product of a sequence of unbounded value.
	
	The Truncated Relative Importance Sampling (TRIS), expressed as $\mu^{tr}_t = min (c, \mu_t^r)$, can stabilise the training as it truncates the min value of the ratio to $c$, which introduces a lower bound to RIS.
	
	\paragraph{Penalty in Policy Gradient.}
	A method to further reduce the variance of TRIS and stabilize the training is encouraging the learnt policy to be close to the behaviour policy~\cite{wu2019behavior}. We can penalise the KL divergence in the value function. It is also related to the KL-control problem in which a KL value penalty is introduced in the value function. Formally, if $\tau=\{a_1, a_2, ..., a_{t-1}\}$, we define the penalised loss objective as:
	\begin{equation}
	\nabla_{\theta}L(\theta)= \mathbb{E}_{\pi^b} \Big \{(\text{CIDEr}^s - \text{CIDEr}\textsuperscript{$\wedge$}) (\prod_{t=0}^T{\mu^{tr}_t}) \nabla_{\theta} log\pi_{\theta}(a_t|s_t)  \Big \} - \beta D[\pi(\tau) || b(\tau)],
	\end{equation}
	where $\mu_t$ is the RIS ratio, and $D$
	is a divergence function between distributions over actions such as KL divergence. This formulation can penalise the behaviour of the target policy being divergent from the behaviour policy. As $D_{kl}[q(x)||p(x)] = \sum_{x} q(x)(\log q(x) - \log p(x))$. Then the loss objective is equivalent to the following expression at the action level:
	\begin{equation}
	\nabla_{\theta}L(\theta)= \mathbb{E}_{\pi^b} \Big \{  [(\text{CIDEr}^s - \text{CIDEr}\textsuperscript{$\wedge$}) + \beta (\log b(a_t|s_t) - \log \pi(a_t|s_t)) ] (\prod_{t=0}^T\mu^{tr}_t) \nabla_{\theta} log\pi_{\theta}(a_t|s_t)  \Big \},
	\end{equation}
	where the term $b(a_t|s_t)$ rewards the model for choosing the action that have a high probability of the prior (behaviour) policy. $-\log \pi(a|s)$ is the entropy regularisation~\cite{ahmed2018understanding}, which is very important in RL for efficient exploring. $\beta$ is the coefficient weight to control the contribution of the penalty term.
	\paragraph{The Rationale of Combining TRIS with KL-control.}
	A fundamental issue of IS is the choice of the importance function, which, in our case, is the behaviour policy $\pi^b$.
	$\pi(E=e) = \frac{1}{M}\sum_{i=1}^{M}\frac{\pi(h_i,e)}{\pi^b(h_i)}$ where $h_i$ is the instantiation of variables $H$ in the $i^{th}$ samples, $e$ is the observed variable. The optimal importance function is when $\pi^b=\pi(H|e)$, which is proportional to $\pi(H, e)$ and lead to zero variance of the IS. In practice, the optimal is not easy to sample. Hence, many researchers are seeking methods to reduce the variance of IS.
	
	While the optimal is hard to find, the KL-divergence between the two distributions can significantly affect the variance of IS, which is proved in~\cite{wexler2012importance}: Let $\pi^{b_1}$ and $\pi^{b_2}$ be two importance functions, and the $D(\pi^{b_1}||\pi) - D(\pi^{b_2}||\pi) = d>\ln c >0$ where $D$ is the KL-divergence, then $\frac{E_{b_1}[Var(\pi/\pi^b)]}{E_{b_2}[Var(\pi/\pi^b)]}
	\geq \frac{e^{2d}}{c^2}$. $Var$ indicates the variance. Accordingly, even a small change in KL-divergence could exponentially alter the variance of IS and RIS. Consequently, we can penalise the target policy when it is divergent from the behaviour policy, to further reduce the variance of TRIS.
	
	\subsection{The Model Structure}
\begin{figure}[t]
		\centering
		\includegraphics[height=8cm, width=\linewidth]{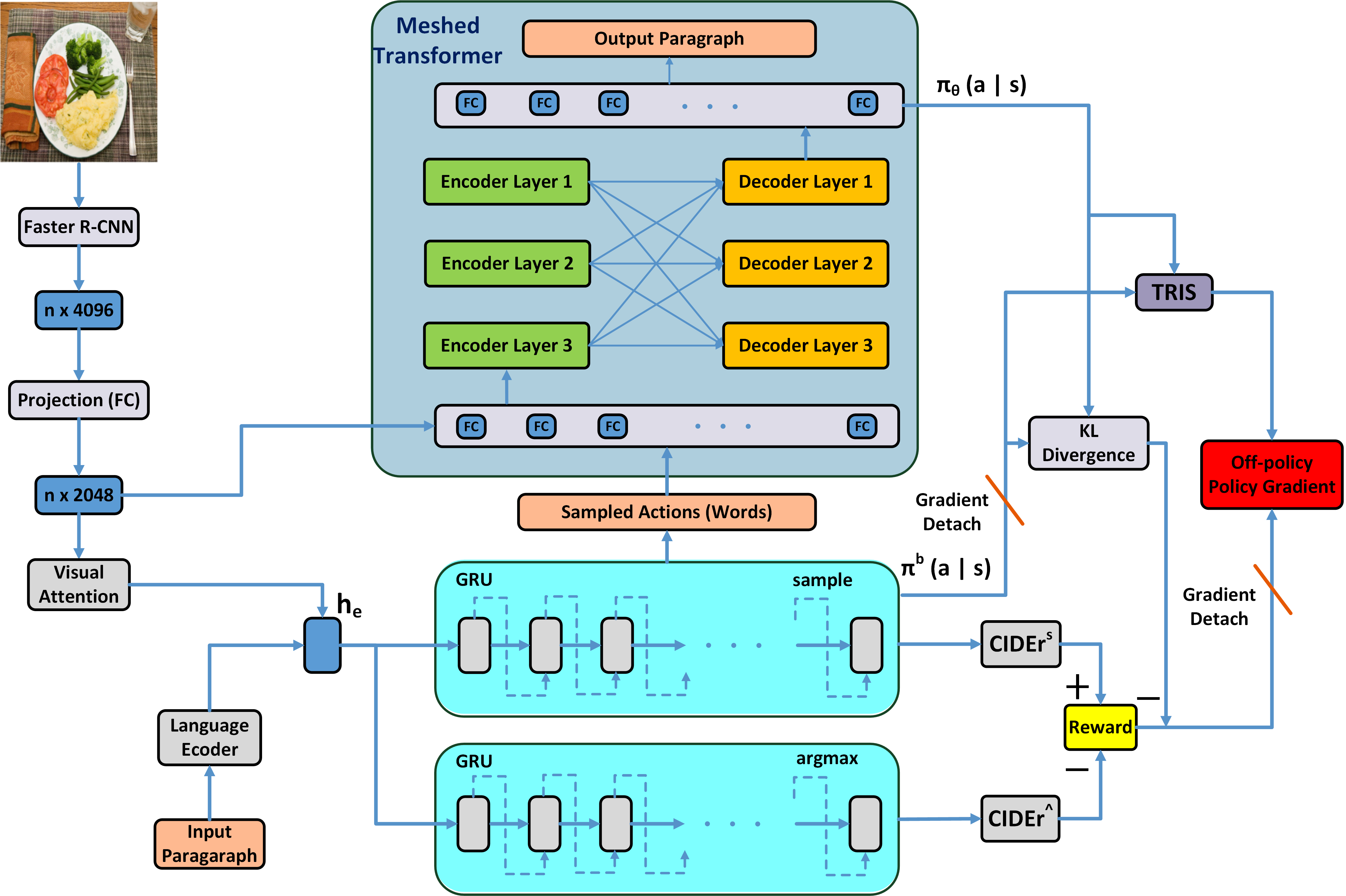}    \caption{The off-policy self-critical for visual paragraph generation: The image is first input to a Faster R-CNN model~\cite{ren2015faster} to extracting $n$ region features, each is with a dimension of 4096. The features are forwarded to Transformer to perform training, after Fully-connected (FC) transforming. Meanwhile, the input paragraph is encoded via a GRU encoder to a hidden vector. The hidden vector, along with the visual features, are subsequently input to a GRU decoder to perform Multinomial Sampling. The sampled words are then forwarded to Transformer to obtain the action probabilities. The self-critical reward obtained from the GRU decoder is formulated with a KL penalty term, which is to reduce the variance of TRIS used for re-weighting the probabilities. \textit{Best Viewed in Colour.}}
		\label{img:system}
		\vspace{-0.5cm}
	\end{figure}

	The agent we utilise is the meshed Transformer~\cite{cornia2019m}, which shows state-of-the-art performance in image captioning. We use a GRU-based image-guided language auto-encoder as the behaviour policy, as shown in Figure~\ref{img:system}. The input paragraph (ground-truth paragraph) is encoded via a GRU-based language encoder to a hidden vector $h_e$, with a size of $batch\times M$. Then we feed the region image features extracted from a pre-trained Faster R-CNN model, denoted as $F = \{f_1, f_2, ... f_K\}$, each item with a size of $batch\times N$ to a visual attention module~\cite{xu2015show} in every time step of the language decoder. Hence the input to the language decoder (a GRU model) at each time step $t$ is expressed as:
	\begin{equation}
	\begin{split}
	& e_{ti} = concat(F, h_e, h_{t})* W_\alpha \\
	& \alpha_{ti} = \frac{exp(e_{ti})}{\sum_{k=1}^K exp(e_{tk})} \\
	& I_t = \sum_{i = 1}^K (\alpha_{ti} * f_i) \\
	& h_t =
	\left\{
	\begin{split}
	& h_e \  \ \text{if} \ t=0 \\
	& \text{GRU}(I_t, h_t) \ \ \text{if} \  t>0,
	\end{split}
	\right .
	\end{split}
	\end{equation}
	where $W_\alpha \in {R}^{L\times1}$ and $L= 2*M+N$.
	The hidden vector of the language decoder is initialised with $h_e$. The language auto-encoder can be considered as image-guided. $h_t$ is then decoded to paragraph. We use the language auto-encoder as the behaviour policy to explore in the environment. To approximate the off-policy learning, TRIS and a KL-divergence penalty are utilised in training.
	\subsection{Training Algorithm}
	We first train an image-guided language auto-encoder using the image-paragraph pairs provided with the dataset, which is then used as a behaviour policy. Transformer~\cite{cornia2019m} is pre-trained using the standard MLE learning scheme on the dataset. Then we treat Transformer model as the target policy, and start the off-policy Policy Gradient training described previously. When training the model under RL, the total loss objective is a combination of MLE loss and RL loss, expressed as:
	\begin{equation}
	\begin{split}
	& Loss_{MLE}(\theta) = -\sum_{t=0}^T log(\pi_\theta(a_t|a_{0:t-1}, I_F))               \\
	& Loss_{total}(\theta) = (1-\alpha)*Loss_{MLE}(\theta) - \alpha* L(\theta), \\
	\end{split}
	\end{equation}
	where the MLE loss is to minimise the negative log probabilities of the generated word token given previous generated work tokens.

	\section{Experiments}
	We conduct the experiments on two use cases of the off-policy self-critical for image-based language generation: visual paragraph generation and image captioning. The merits of our algorithm are mainly in tackling long sequence generation for Transformer models in, e.g., visual paragraph generation. Image captioning is to generate a caption for a given image, which can be combined with on-policy self-critical~\cite{cornia2019m}. Nevertheless, we apply the proposed method on image captioning as well.
	\subsection{Visual Paragraph Generation}
	\paragraph{Implementation Details.}
	We experiment on the Stanford Visual Paragraph dataset~\cite{krause2017hierarchical}. In this dataset, each image contains one paragraph. The training, validation and testing sets contain 14,575, 2487 and 2489 images, respectively. We evaluate the BLEU, METEOR, ROUGE-L and CIDEr scores for the generated paragraphs. For MLE baseline, we train the model for 40 epochs. For our off-policy self-critical algorithm, we further train the model for 8 epochs using a combination of off-policy RL and MLE. We use early stopping on CIDEr score to choose the best model for every one epoch. The learning rate is set as 4e-4 for MLE training, and 4e-5 for our off-policy self-critical training. We use Adam optimiser~\cite{kingma2014adam} with stochastic back-propagation. The batch size is set as 20. Our experiments are conducted using Pytorch 1.2.0 and with a server equipped with an NVIDIA 2080-Ti GPU.

	\paragraph{Ablation Studies.}
	Firstly, we set two kinds of behaviour policies, the visual attention-based captioning model~\cite{xu2015show} and our image-guided language auto-encoder, which are shown in Table~\ref{policy}. The attention model yields poorer performance than our auto-encoder as we include the language information in our auto-encoder.
	
	The impact of such different behaviour policy on the target policy is not that obvious, as revealed in Table~\ref{policy_performance}. This phenomenon shows that: (1). the behaviour policy is only applied in the exploration of RL, which, in theory, does not affect the target policy. (2). Behaviour policy that selects better action, can have a better impact on the target policy as the reward tends to be more positive.
	
	RIS can reduce the variance of IS via a simple technique of linear transformation. As the reduced variance leads to more stable training, the performance can be raised, as shown in Table~\ref{ris}. TRIS can further boost the performance as it additionally introduces a lower bound of the RIS ratio. This lower bound guarantees that the $\prod_{t=0}^T{\mu^{tr}_t}$ is bigger than zero mostly, leading to more effective training.
	
	The KL-control technique described previously can penalise the target policy when it is divergent from the behaviour policy, thus can reduce the variance of IS. The results are shown in Table~\ref{ris} and Table~\ref{ris_kl}. The TRIS with KL-control can increase the final performance of the target policy.
	
	We study the value of $c$ in TRIS, as presented in Table~\ref{c}. A suitable $c$ is critical in maintaining the performance as it directly affects the TRIS ratio. $c = 0.96$ yields the best results.
	
	The coefficient $\alpha$ also has an impact on the performance, $\alpha = 0.5$ can make a right balance between supervised learning and off-policy RL learning, as shown in Table~\ref{all}.
	
	\begin{figure}
		\centering
		\begin{subfigure}[t]{0.24\textwidth}
			\centering
			\includegraphics[height = 2.7cm, width=\textwidth]{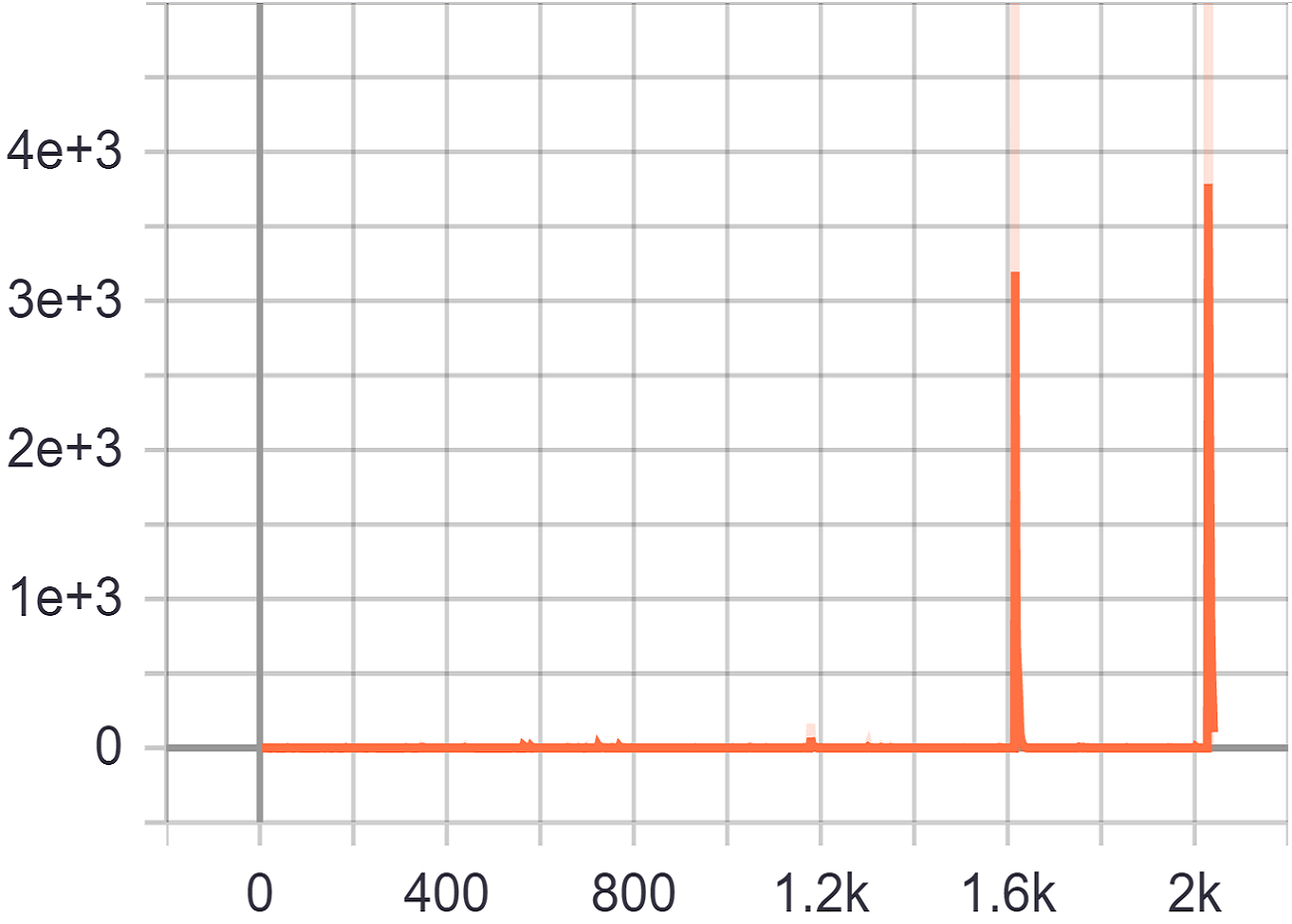}
			\caption{IS ratio.}
			\label{}
		\end{subfigure}
		\hfill
		\begin{subfigure}[t]{0.24\textwidth}
			\centering
			\includegraphics[height = 2.7cm, width=\textwidth]{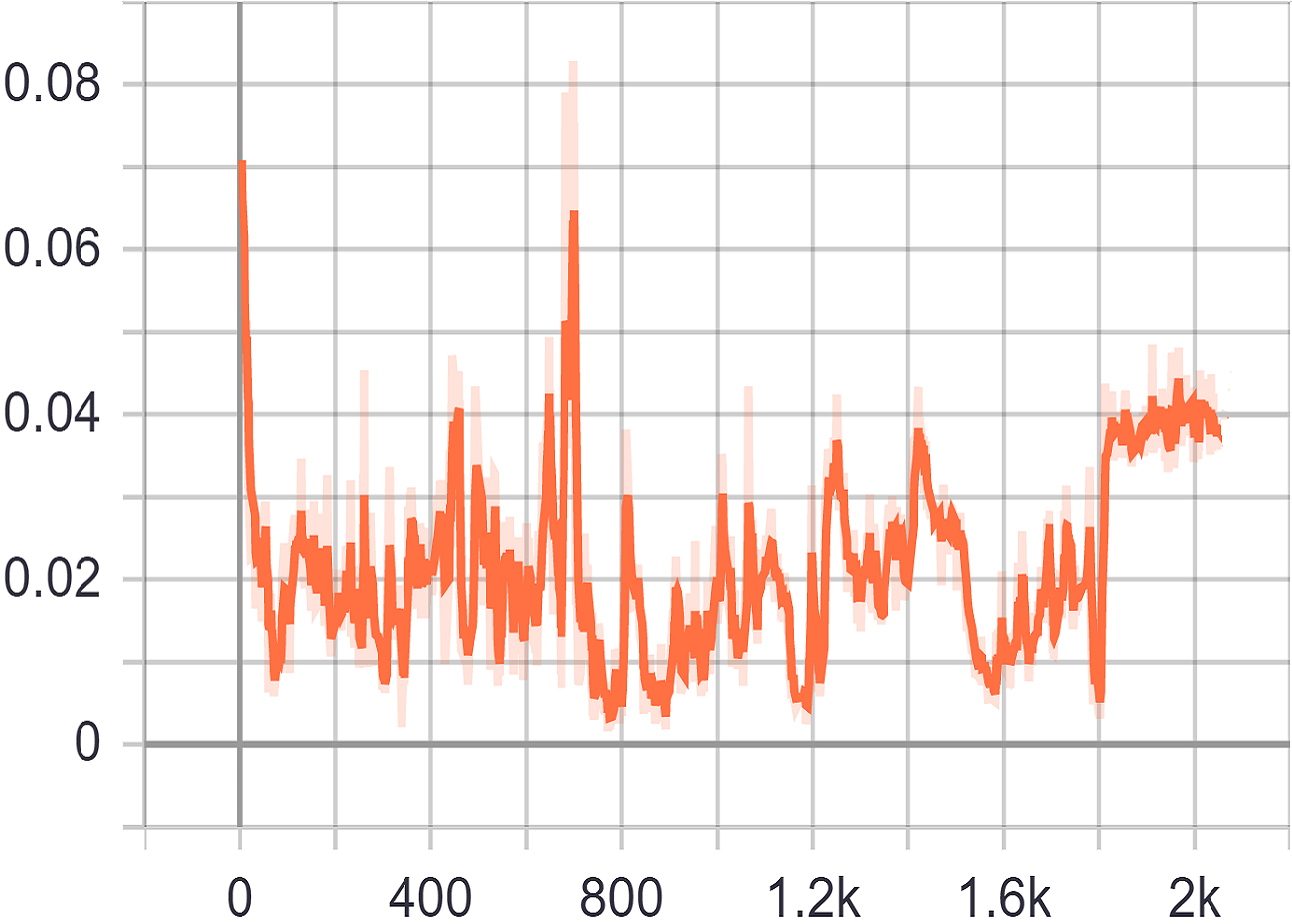}
			\caption{RIS ratio.}
			\label{}
		\end{subfigure}
		\hfill
		\begin{subfigure}[t]{0.24\textwidth}
			\centering
			\includegraphics[height = 2.7cm, width=\textwidth]{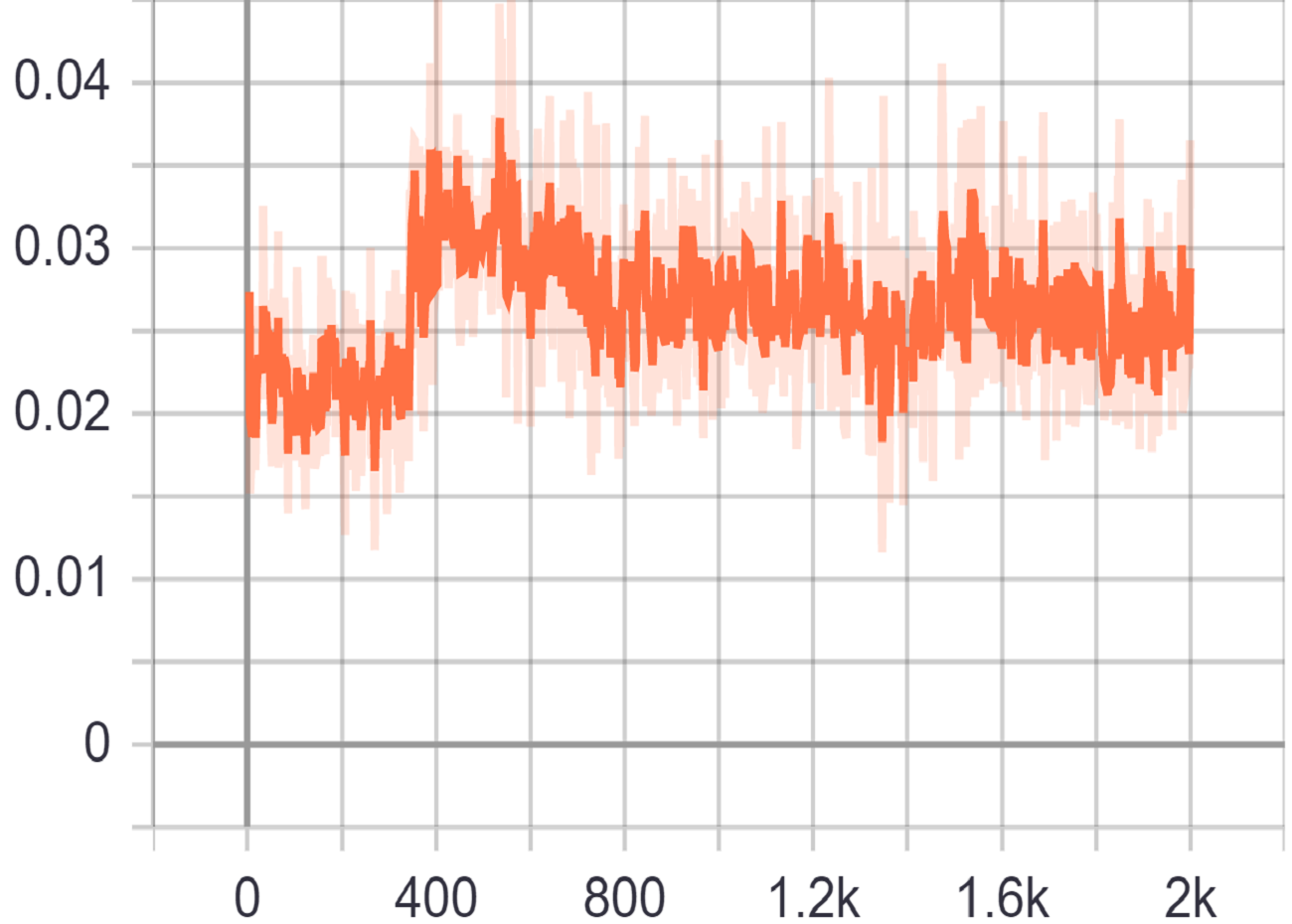}
			\caption{RIS ratio with KL penalty.}
			\label{}
		\end{subfigure}
		\begin{subfigure}[t]{0.24\textwidth}
			\centering
			\includegraphics[height = 2.7cm,width=\textwidth]{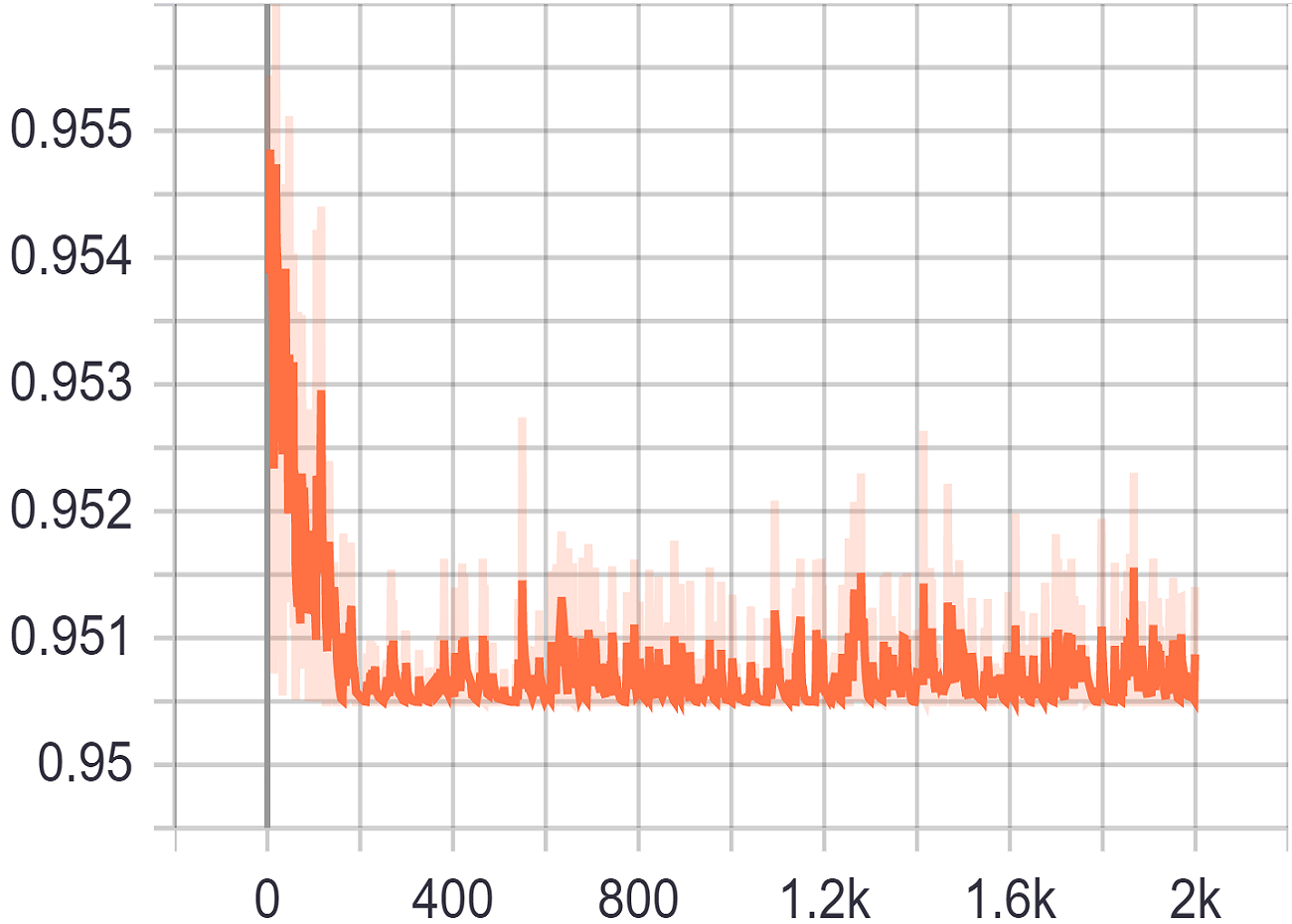}
			\caption{TRIS ratio with KL penalty.}
			\label{}
		\end{subfigure}
		\caption{The IS ratio of different schemes. The X-axis is the iterations while the Y-axis is the ratio. We see an obvious impact of TRIS and KL penalty on the ratio value.}
		\label{img:is_r}
		\vspace{-0.2cm}
	\end{figure}
	
	We plot the IS ratio curves of training versus the iteration. We run the RL training for 2000 iterations, with a batch size of 20, which can be seen in Figure~\ref{img:is_r}. The IS ratio leads to a very high value (more than 3000) in around 1600 and 2000 iterations, which is not bounded. RIS with a relative ratio of 0.5 can significantly reduce the variance, making the value of IS ratio below 0.07, which shows critical contrast with the IS ratio. The KL-control can further reduce the variance of the RIS ratio, limiting the RIS ratio below 0.05. The TRIS introduces a lower bound of 0.95 to the ratio, leading to stable training.

	\begin{table*}[!t]
		\caption{The performance of behaviour policies.}
		\resizebox{\linewidth}{!}{
			\centering
			\begin{tabular}{cccccccc}
				\toprule
				Methods  & BLEU-1 & BLEU-2 & BLEU-3 & BLEU-4 & METEOR & ROUGE-L&  CIDEr \\
				\midrule
				behaviour Policy~\cite{xu2015show} & 22.4  & 10.2 & 4.3 & 1.7 & 9.5  & 24.4 & 7.0 \\
				Our behaviour Policy & \textbf{46.3} & \textbf{30.8} & \textbf{20.9} & \textbf{14.5} & \textbf{18.0} & \textbf{41.5} & \textbf{66.7} \\
				\bottomrule
			\end{tabular}
		}    \label{policy}
	\end{table*}
	
	\begin{table*}[!t]
		\caption{The impact of behaviour policies on the performance of the target policy.}
		\resizebox{\linewidth}{!}{
			\centering
			\begin{tabular}{cccccccc}
				\toprule
				Methods  & BLEU-1 & BLEU-2 & BLEU-3 & BLEU-4 & METEOR & ROUGE-L&  CIDEr \\
				\midrule
				off-policy with behaviour Policy~\cite{xu2015show} & 37.8 & 22.1 & 13.0 & 7.6 &  14.9 & 29.2& 14.1 \\
				off-policy with our behaviour Policy  & \textbf{41.9} & \textbf{24.8} & \textbf{14.8} & \textbf{8.9} & \textbf{16.6} & \textbf{29.8} & \textbf{19.0}  \\
				\bottomrule
			\end{tabular}
		}    \label{policy_performance}
	\end{table*}

	\begin{table*}[!t]
		\caption{The impact of TRIS on the performance of the target policy.}
		\resizebox{\linewidth}{!}{
			\centering
			\begin{tabular}{cccccccc}
				\toprule
				Methods  & BLEU-1 & BLEU-2 & BLEU-3 & BLEU-4 & METEOR & ROUGE-L&  CIDEr \\
				\midrule
				IS, $\alpha = 0.5$ & 41.9& 24.8 & 14.8 & 8.9 & 16.6 & 29.8 & 19.0 \\
				IS + KL, $\beta = 0.05, \alpha = 0.5 $ & 42.1 & 24.2 & 14.1 & 9.2& 16.5& 28.2&  16.9\\
				RIS + KL, $\beta = 0.05, \alpha = 0.5$  & \textbf{43.1} & 25.5 & 15.2 & 9.0 & 16.9 & 29.5 & 20.0 \\
				TRIS + KL, $\beta = 0.05, \alpha = 0.5 $ & 42.7& \textbf{25.7} & \textbf{15.5} & \textbf{9.4} & \textbf{16.9} &
				\textbf{30.2} & \textbf{20.9} \\
				\bottomrule
			\end{tabular}
		}    \label{ris}
	\end{table*}

	\begin{table*}[!t]
		\caption{The impact of coefficient weight $\beta$ of KL-control on the performance on the target policy.}
		\resizebox{\linewidth}{!}{
			\centering
			\begin{tabular}{cccccccc}
				\toprule
				Methods  & BLEU-1 & BLEU-2 & BLEU-3 & BLEU-4 & METEOR & ROUGE-L&  CIDEr \\
				\midrule
				RIS+KL, $\beta = 0.2, \alpha = 1.0$  & 42.0 & 24.4 & 14.4 & 8.6 & 16.5 & 29.0 &    19.9 \\
				RIS+KL, $\beta = 0.1, \alpha = 1.0$ & 42.9 & 25.4 & 15.1 & 8.9 & 16.7 & 29.7 & 20.2 \\
				RIS+KL, $\beta = 0.05, \alpha = 1.0$ & \textbf{42.5} & \textbf{25.4} & \textbf{15.3} & \textbf{9.2} & \textbf{16.7} & \textbf{30.1} & 19.5 \\
				\bottomrule
			\end{tabular}
		}  \label{ris_kl}
	\end{table*}
	\begin{table*}[!t]
		\caption{The impact of the coefficient of the off-policy policy gradient on the performance.}
		\resizebox{\linewidth}{!}{
			\centering
			\begin{tabular}{cccccccc}
				\toprule
				Methods  & BLEU-1 & BLEU-2 & BLEU-3 & BLEU-4 & METEOR & ROUGE-L&  CIDEr \\
				\midrule
				RIS+KL, $\alpha = 0.2$ & 42.3 & 24.8 & 14.6 & 8.5 & 16.5 & 29.1 & 18.8 \\
				RIS+KL, $\alpha = 0.5$  & \textbf{43.1} & \textbf{25.5} & 15.2 & 9.0 & \textbf{16.9} & 29.5 & 20.0 \\
				RIS+KL, $\alpha = 0.8$  & 42.3 & 25.1 & 15.0 & 8.9 & 16.6 & 29.6 & \textbf{20.2} \\
				RIS+KL, $\alpha = 1.0$  & 42.5 & 25.4 & \textbf{15.3} & \textbf{9.2} & 16.7 & \textbf{30.1} & 19.5 \\
				\bottomrule
			\end{tabular}
		}    \label{all}
	\end{table*}
	
	\begin{table*}[!t]
		\caption{The impact of the truncated value $c$ on the performance of TRIS.}
		\resizebox{\linewidth}{!}{
			\centering
			\begin{tabular}{cccccccc}
				\toprule
				Methods  & BLEU-1 & BLEU-2 & BLEU-3 & BLEU-4 & METEOR & ROUGE-L&  CIDEr \\
				\midrule
				TRIS+KL, $c = 0.96$ & \textbf{44.2} & \textbf{26.8} & \textbf{16.2} & \textbf{9.8} & \textbf{17.2} & \textbf{30.8} & 20.6 \\
				TRIS+KL, $c = 0.95$ & {42.7} & {25.7} & {15.5} & {9.4} & {16.9} &
				{30.2} & \textbf{20.9} \\
				TRIS+KL, $c = 0.85$ & 41.6 & 24.7 & 14.7 & 8.7 & 16.5 & 29.5 & 19.9 \\
				\bottomrule
			\end{tabular}
		}    \label{c}
	\end{table*}

	\paragraph{Comparison with the State-of-the-art.}
	The comparison of our scheme and the current leading methods are shown in Table~\ref{result}. We achieve state-of-the-art results by using our algorithms with a Transformer optimised on CIDEr. The achieved results even significantly outperform the human's annotations on BLEU scores. The CIDEr score is also state-of-the-art.
	
	\begin{table*}[!t]
		\caption{The Performance Comparison with the State-of-the-art Methods on the Stanford Visual Paragraph Dataset.}
		\vspace{0.1cm}
		\resizebox{\linewidth}{!}{
			\renewcommand\arraystretch{1}
			\centering
			\begin{tabular}{cccccccc}
				\toprule
				Category &    Methods  & BLEU-1 & BLEU-2 & BLEU-3 & BLEU-4 & METEOR & CIDEr \\
				\hline
				\multirow{6}{*}{Flat Models} &
				Sentence-Concat~\cite{krause2017hierarchical} & 31.1 & 15.1 & 7.6 & 4.0 & 12.1 & 6.8 \\
				&    Template~\cite{krause2017hierarchical} & 37.5 & 21.0 & 12.0 & 7.4 & 14.3 & 12.2 \\
				&        Image-Flat~\cite{krause2017hierarchical} & 34.0 & 19.1 & 12.2 & 7.7 & 12.8 & 11.1  \\
				&        Top-down Attention~\cite{anderson2018bottom} &32.8 &19.0 &11.4 &6.9 &12.9 &13.7 \\
				&        self-critical~\cite{rennie2017self}  &29.7& 16.5 &9.7 &5.9& 13.6& 13.8 \\
				&        DAM-Att~\cite{wang2018look} &35.0 &20.2 &11.7 &6.6 &13.9 &17.3 \\
				& \textbf{Meshed Transformer + MLE~\cite{cornia2019m}} & 37.5 & 22.3 & 13.7 & 8.4 & 15.4 & 16.1 \\
				\midrule
				\multirow{4}{*}{Hierarchical Models} &
				Regions-Hierarchical~\cite{krause2017hierarchical} & 41.9 & 24.1 & 14.2 & 8.7 & 16.0 & 13.5  \\
				&        RTT-GAN~\cite{liang2017recurrent} & 42.0 & 24.9 & {14.9} & 9.0 &  17.1 & 16.9  \\
				&        Diverse (VAE)~\cite{chatterjee2018diverse} &  42.4  & {25.6} & {15.2} & {9.4} & {18.6} & {20.9} \\
				& {ParaCNN~\cite{yan2020paracnn}} & 42.0 & 25.0 & 14.9 &8.8& 17.0 & 20.4\\
				\midrule
				\multirow{2}{*}{Ours}
				
				& \textbf{Meshed Transformer~\cite{cornia2019m} + off-policy (c = 0.95)}   & {42.7} & {25.7} & {15.5} & {9.4} & {16.9} & {20.9} \\
				
				& \textbf{Meshed Transformer~\cite{cornia2019m} + off-policy (c = 0.96)}   & \textbf{44.2} & \textbf{26.8} & \textbf{16.2} & \textbf{9.8} & {17.2}  & 20.6 \\
				
				\midrule
				Human~\cite{krause2017hierarchical} & Annotations  &      42.9 & 25.7 & 15.6 & 9.7  & \textbf{19.2}  & \textbf{28.6} \\
				\bottomrule
			\end{tabular}
		}
		\label{result}
	\end{table*}%

	\subsection{Extending to the Convolutional Model and Image Captioning}
	Convolutional captioning~\cite{aneja2018convolutional} has a similar parallel computing feature to Transformer. The sentence needs to be generated is shorter, requiring relatively less GPU computing resources. Nevertheless, we test our off-policy learning on the convolutional image captioning task. Following the practice of~\cite{aneja2018convolutional}, We experiment on MS-COCO dataset~\cite{lin2014microsoft} under the `Karpathy' split and report results, which are presented in Table~\ref{ip-result}. We follow the training protocol of the paper~\cite{aneja2018convolutional} for the baseline. We further train the model for 5 epochs using our off-policy self-critical algorithm. Our method improves the convolutional captioning in almost every metric of language evaluation. Notably, the CIDEr is significantly enhanced as our off-policy RL is optimised towards the CIDEr score.

	\begin{table*}[!t]
		\caption{The impact of off-policy RL on convolutional image captioning.}
		\resizebox{\linewidth}{!}{
			\centering
			\begin{tabular}{cccccccc}
				\toprule
				Methods  & BLEU-1 & BLEU-2 & BLEU-3 & BLEU-4 & METEOR & ROUGE-L&  CIDEr \\
				\midrule
				Conv Captioning~\cite{aneja2018convolutional}  &71.0 &   53.6  &  38.9   &27.9  &24.0 &  {51.9}  & 88.1 \\
				Our off-policy learning & {70.9} & \textbf{53.8} &
				\textbf{39.3} & \textbf{28.3} & \textbf{24.2} & \textbf{52.0}&  \textbf{90.0}\\
				\bottomrule
			\end{tabular}
		}    \label{ip-result}
	\end{table*}

	\section{Conclusions}
	Transformer and Convolution-based seq-to-seq model is hard to perform on-line RL optimisation in visual paragraph generation as the computing resources required for a such models are beyond current equipments. Hence, we propose an off-policy self-critical algorithm in which a GRU-based model is set as the behaviour policy to perform sampling in RL, whose sampling is much more efficient. To better approximate this off-policy RL, both TRIS and a KL-divergence penalty term are applied in the off-policy RL to reduce the high variance of the IS approximation. As Transformer is empowered with RL learning capability, we achieve state-of-the-art results on visual paragraph generation and also improved results on image captioning.

	\section*{Broader Impact}
	In this paper, we introduce an off-policy self-critical sequence training algorithm, especially targeting on Transformer-like models in visual paragraph generation, enabling the feasibility of the combination of these advanced models and reinforcement learning (RL).
	
	Usually, we can consider the language generation task as a sequential decision-making process. At each time step, the agent selects a word from a pre-defined vocabulary until the whole sentence or paragraph is generated.  Previous studies usually make the agent perform on-policy. However, the off-policy can prevent the agent from real exploration and is sample efficient. Transformer is especially computing-expensive in on-policy exploration, leading to the in-feasibility of the on-policy RL for Transformer. One of the impact is that the proposed algorithm can not only directly save computing resource by preventing Transformer from real exploration but also show the possibilities of the off-policy RL in language generation tasks.
	
	This research is also a test on how the off-policy RL performs in large action space problems. The off-policy is usually rooted in a value-based RL algorithm where the action space is small. Instead, we propose a policy gradient method, without Temporal Difference (TD) bootstrapping, to directly transfer the Monte-Carlo experience of the behaviour policy to the target policy. Mostly, the policy gradient is better behaved when combined with function approximations, while the TD method is more readily applied in off-policy learning. The main obstacle is the high variance in the off-policy estimation of the policy gradient. We show that it is feasible to formulate and apply the off-policy policy gradient if we handle the variance properly.
	
	A drawback of the proposed algorithm is that we might need to introduce another RNN-based model as the behaviour policy, which increases the number of the parameters of the models. Hence, further research can make efforts on how to reduce the training models' complexity while achieving the same effects of the off-policy RL.
	
	In summary, this research can help existing natural language processing (NLP) models like Transformer to perform off-policy RL learning, which is a novel way of the training of Transformer. This research will also provide insights for other RL learning scheme, for instance, actor-critic learning, in various NLP tasks.

	
	
	\bibliographystyle{unsrt}
	\bibliography{bib}
\end{document}